\documentclass[]{spie}  

\usepackage{amsmath,amsfonts,amssymb}
\usepackage{graphicx}
\usepackage[colorlinks=true, allcolors=blue]{hyperref}
\usepackage[numbers,sort&compress]{natbib}
\usepackage{booktabs}
\usepackage{subcaption}
\usepackage{threeparttable} 

\title{Benchmarking and Enhancing Surgical Phase Recognition Models for Robotic-Assisted Esophagectomy}

\author[a]{Yiping Li}

\author[b]{Romy van Jaarsveld}
\author[a]{Ronald de Jong}
\author[a]{Jasper Bongers}
\author[b]{Gino Kuiper}
\author[b]{Richard van Hillegersberg}
\author[b]{Jelle Ruurda}
\author[a]{Marcel Breeuwer}
\author[a]{Yasmina Al Khalil}
\affil[a]{Eindhoven
University of Technology, Eindhoven, The Netherlands}
\affil[b]{University Medical
Center Utrecht, Utrecht, The Netherlands}

\begin{document} 
\maketitle

\section{Abstract}
Robotic-assisted minimally invasive esophagectomy (RAMIE) is a recognized treatment for esophageal cancer, offering better patient outcomes compared to open surgery and traditional minimally invasive surgery. RAMIE is highly complex, spanning multiple anatomical areas and involving repetitive phases and non-sequential phase transitions. Our goal is to leverage deep learning for surgical phase recognition in RAMIE to provide intraoperative support to surgeons. To achieve this, we have developed a new surgical phase recognition dataset comprising 27 videos. Using this dataset, we conducted a comparative analysis of state-of-the-art surgical phase recognition models. To more effectively capture the temporal dynamics of this complex procedure, we developed a novel deep learning model featuring an encoder-decoder structure with causal hierarchical attention, which demonstrates superior performance compared to existing models.
\section{Introduction}
\label{sec:intro}  

Esophageal cancer, known for its high malignancy and poor prognosis, is the 11th most common cancer and ranks 7th in cancer-related mortality worldwide, posing a significant challenge in oncology \cite{bray2024global}. Robotic-assisted minimally invasive esophagectomy (RAMIE) is a recognized treatment procedure for esophageal cancer \cite{fuchs2022robotic}. However, RAMIE is a highly complex surgical procedure that spans multiple anatomical regions, requiring precise navigation and manipulation of various structures. The learning curve for fully minimally invasive RAMIE is substantial, with one study reporting a learning phase of 70 procedures over 55 months \cite{van2018learning}. The application of machine learning to RAMIE procedures is still in its early stages. \citeauthor{den2023deep} \cite{den2023deep} published the first study on key anatomy segmentation in RAMIE procedures with deep learning. \citeauthor{sato2022real} \cite{sato2022real} developed a sophisticated model for laryngeal nerve identification, addressing a critical aspect of patient safety during the procedure. \citeauthor{takeuchi2022automated} \cite{takeuchi2022automated} investigated phase recognition in RAMIE, utilizing their in-house data with TeCNO \cite{czempiel2020tecno} model. More recently, \citeauthor{brandenburg2023active} \cite{brandenburg2023active} demonstrated that active learning can significantly reduce annotation effort while maintaining high machine learning performance for specific surgomic features.

Surgical phase recognition is used in computer-assisted surgery systems to classify different stages of a surgical procedure from video footage. It supports intraoperative decision-making, enhances workflow efficiency, and enables postoperative analysis of surgical phases, surgeon performance evaluation, and identification of problematic phases \cite{maier2020surgical}. In the context of RAMIE, where recognizing crucial anatomical structures remains challenging, we aim to leverage surgical phase recognition to improve contextual understanding and provide preemptive assistance during surgery. Furthermore, as complications often arise from specific high-risk surgical steps \cite{van2019outcomes}, surgical phase recognition is useful in extracting relevant video clips for postoperative analysis. 

With this motivation, our study introduces a dataset designed for RAMIE phase recognition, capturing the complex temporal dynamics inherent in this procedure. We conducted a comparative analysis of various machine learning models applied to this dataset and proposed an enhanced model to improve performance in phase recognition tasks. Our objective is to establish a robust foundation for future model development in this area. Surgical phase recognition serves as an initial step for more advanced, data-driven analyses of surgical procedures. By contributing to the evolving landscape of surgical data science for RAMIE, we seek to enhance surgical training, optimize workflows, and improve patient outcomes in esophageal cancer treatment.

\section{METHODS}
\subsection{Data}
\subsubsection{RAMIE Dataset}
This study utilizes a specialized database of 27 randomly selected Robot-Assisted Minimally Invasive Esophagectomy (RAMIE) recordings obtained from the surgical recordings repository of the University Medical Center (UMCU), collected between January 2018 and July 2021. While RAMIE typically involves both thoracic and abdominal phases, our research focuses exclusively on the thoracic phase of the procedure due to the complexity of the mediastinum, which contains numerous vital anatomical structures, including the aorta, airways, and laryngeal nerves. We analyzed video footage from the initial camera entry into the thoracic cavity until just before the esophageal division.

\begin{figure} [ht]
\begin{center}
\begin{tabular}{c} 
\includegraphics[height = 4.7 cm]{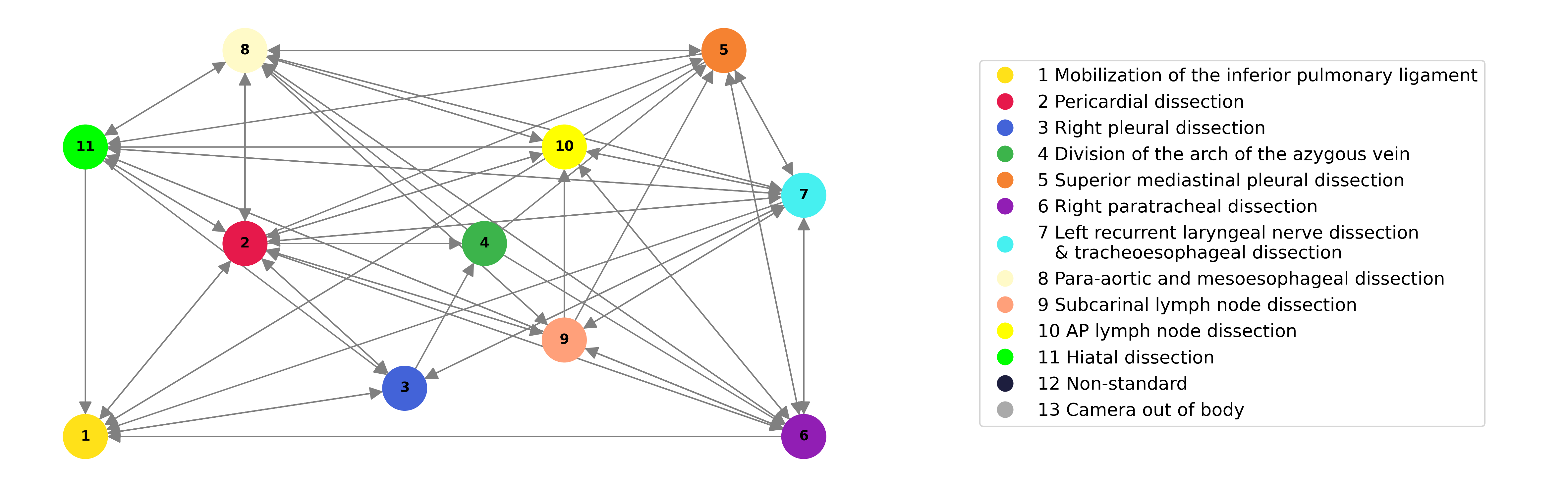}
\end{tabular}
\end{center}
\caption[Phase Labels and transition maps] 
{ \label{phase} 
Schematic representation of RAMIE thoracic phases}
\end{figure}

According to the standardized approach for thoracic dissection in RAMIE outlined by \citeauthor{kingma2020standardized} \cite{kingma2020standardized}, we identified 13 distinct phases within the procedure. This includes 11 surgical phases primarily delineated by anatomical areas, as shown in Figure \ref{phase}, along with additional phases for non-standard actions and camera-out-of-body periods. Non-standard actions include transitions involving excessive camera movements, encircling of the esophagus to connect anatomical areas, and abnormal events such as major bleeding or irrigation. Variability in phase sequence is significant across cases, given the surgeon's operating habits and patient anatomy. While the annotated phases should typically follow a standard numerical order, interruptions may occur when one anatomical plane is entered during a non-corresponding phase. Numbers and arrows in the figure indicate the typical progression and possible transitions between phases in this dataset.

\begin{figure} [htbp]
\begin{center}
\begin{tabular}{c} 
\includegraphics[height = 5.5 cm]{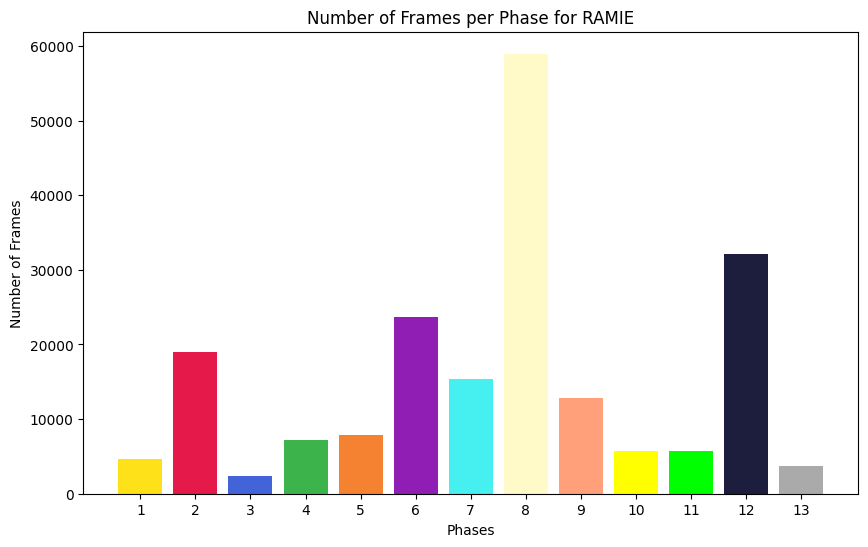}
\end{tabular}
\end{center}
\caption[Phase Labels and transition maps] 
{ \label{numberofframes} 
Number of frames per phase in RAMIE dataset}
\end{figure}

The surgical phases in these videos were annotated by a PhD student in biomedical engineering, guided by a medical PhD student and an expert surgeon. Video labelling was performed at 25 frames per second (fps). The dataset was divided into 14 videos for training, 4 for validation, and 9 for testing. Following current research practices, all machine learning models in this study were trained at 1 fps, resulting in 105,387 frames for training, 27,249 frames for validation, and 66,596 frames for testing. Figure \ref{numberofframes} shows the number of frames for each phase.

\subsubsection{AutoLaparo Dataset}
In addition to evaluating our proposed model on the RAMIE dataset, we conducted experiments using the publicly available AutoLaparo dataset \cite{wang2022autolaparo}, a widely used benchmark in this domain. This dataset comprises full-length videos of complete hysterectomy procedures with annotations for seven distinct phases: \textit{Preparation, Dividing Ligament and Peritoneum, Dividing Uterine Vessels and Ligament, Transecting the Vagina, Specimen Removal, Suturing, and Washing}. The sequence of Phase 2 and Phase 3 may differ based on the surgeon’s operating habits. Annotations for AutoLaparo were performed by a senior gynecologist with over thirty years of clinical experience, supported by a specialist with three years of hysterectomy experience. The dataset includes 21 videos, divided into training (10 videos, 40,211 frames), validation (4 videos, 12,056 frames), and testing (9 videos, 12,056 frames).

\subsection{Surgical Phase Recognition Models for Benchmarking}
We selected four state-of-the-art surgical phase recognition models for benchmarking: SV-RCNet \cite{jin2017sv}, TMRNet 
 \cite{jin2021temporal}, TeCNO \cite{czempiel2020tecno}, and Trans-SVNet \cite{gao2021trans} based on their demonstrated effectiveness in the AutoLaparo \cite{wang2022autolaparo} and Cholec80 \cite{twinanda2016endonet} datasets, with both being widely used benchmarks in the field of surgical video analysis. We implemented these methods using the open-source code provided by the original authors, maintaining all original settings in the code to ensure consistency and comparability.

SV-RCNet integrates visual and temporal dependencies in an end-to-end architecture, combining a deep ResNet for spatial feature extraction with LSTM networks for capturing temporal dependencies in surgical workflow recognition. TMRNet relates multi-scale temporal patterns using a long-range memory bank and a non-local bank operator, allowing the model to capture both short-term and long-term temporal relationships crucial for understanding complex surgical dynamics. TeCNO exploits temporal modeling with higher temporal resolution and a large receptive field by using a multi-stage temporal convolution network in a causal way, enabling it to capture fine-grained temporal patterns and long-range dependencies across entire surgical videos more effectively than traditional approaches. Trans-SVNet attempts to use Transformer architectures to fuse spatial and temporal embeddings in surgical video analysis, leveraging self-attention mechanisms to potentially capture complex spatial-temporal relationships more effectively than traditional convolutional or recurrent approaches.
\subsection{Proposed Model}
RAMIE is a highly complex surgical procedure characterized by numerous repetitive phases and non-sequential phase transitions. This complexity contrasts with most publicly available surgical phase recognition datasets, which typically feature more sequential processes with limited phase order variations. Consequently, this necessitates more advanced temporal modelling. Inspired by ASformer \cite{yi2021asformer} and its success on the Breakfast \cite{Kuehne12} and 50 Salads \cite{stein2013combining} datasets, which are well-established benchmarks for temporal action segmentation, we identify parallels between these tasks and surgical phase recognition. Both tasks involve capturing intricate temporal dependencies with minimal constraints on phase order. Drawing on ASformer's demonstrated capability to address these challenges, we implemented a causal transformer architecture with an encoder-decoder structure to facilitate sequential information processing for intra-operative surgical phase recognition.

\subsubsection{Model Architecture}
\begin{figure}[ht]
    \centering
    \includegraphics[height=5.3cm]{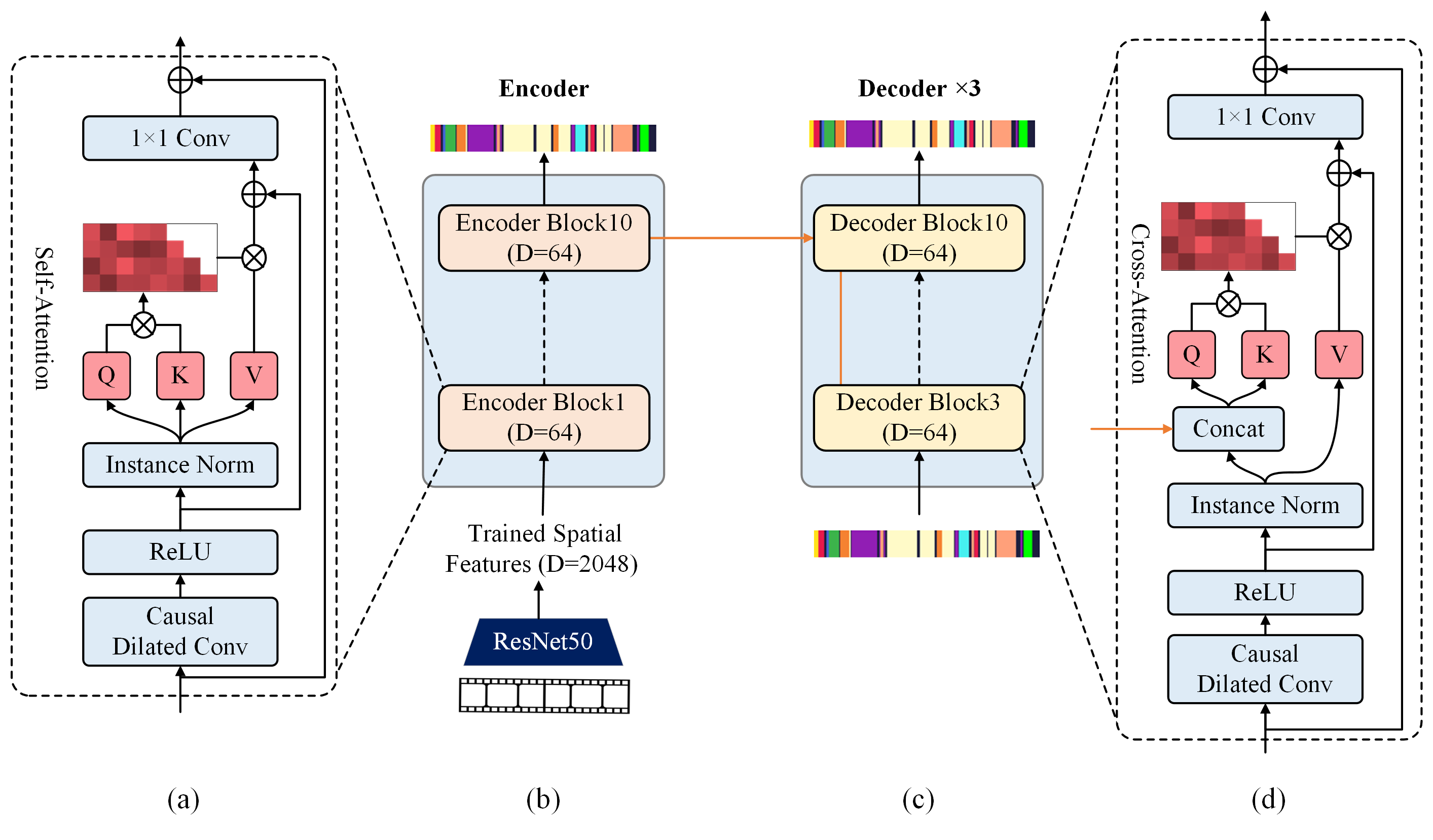}
    \hspace{0.01\textwidth}
    \includegraphics[height=5.8cm]{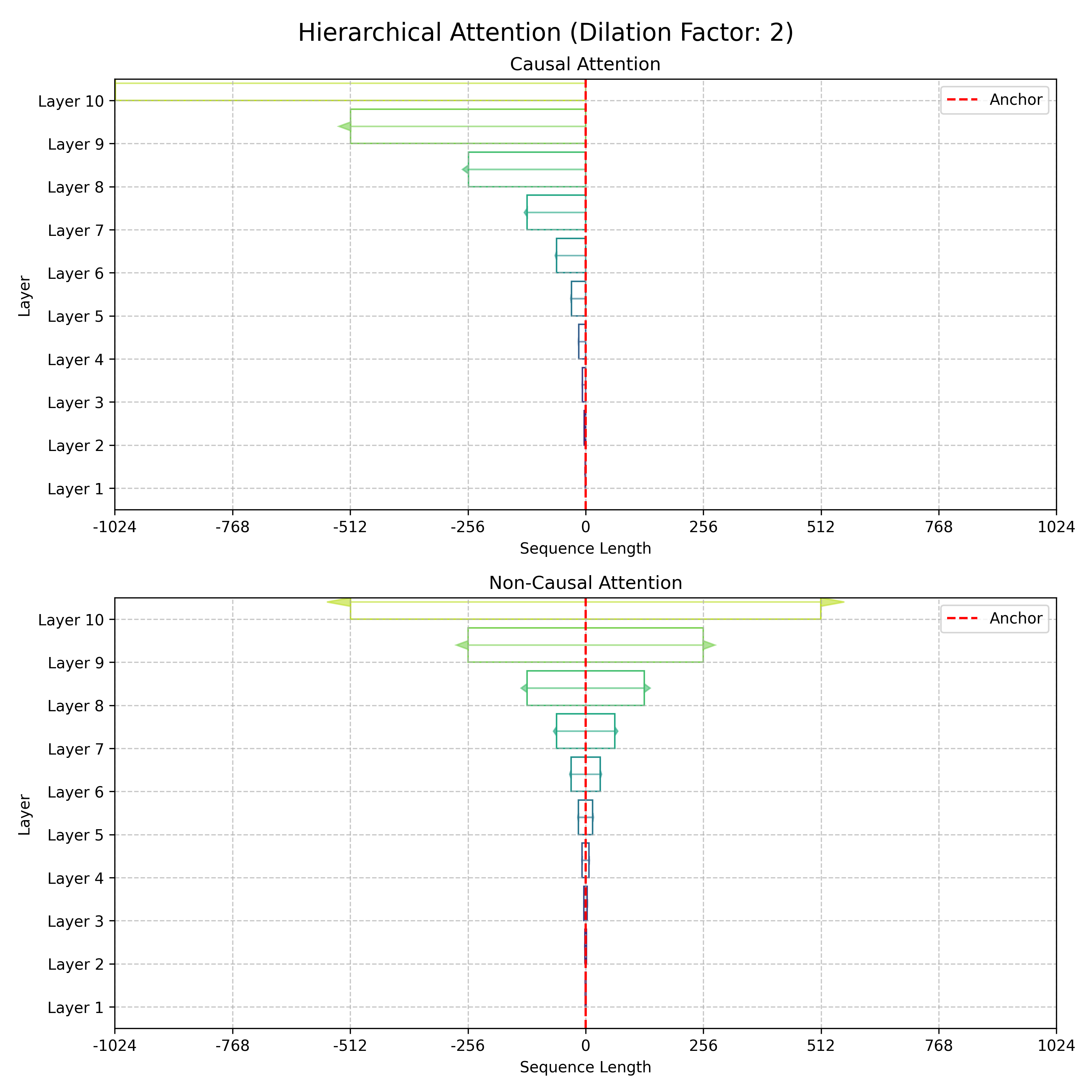}
    \caption{Proposed model architecture (left) and comparison of hierarchical attention in causal and non-causal settings (right), adapted from ASFormer \cite{yi2021asformer}. For each layer $l \in \{1, \ldots, L\}$, the query tensor $Q_l \in \mathbb{R}^{T_l \times d \times h_l}$ and the key tensor $K_l \in \mathbb{R}^{T_l \times d \times 2h_l}$ are defined, where $T_l = \left\lfloor\frac{T_0}{2^{l-1}}\right\rfloor$ is the sequence length, $d$ is the feature dimension, and $h_l = 2^{l-1}$ is the head dimension. A causal mask is applied to ensure that each position can only attend to previous positions in the sequence. The right image illustrates the difference between non-causal (bottom) and causal (top) hierarchical attention resulting from the causal dilated convolution.}
    \label{model}
\end{figure}

Figure \ref{model} illustrates our proposed model, which employs a two-stage training approach: feature extraction followed by temporal modeling. In the first stage, we utilize a ResNet50 model, trained frame-by-frame on phase labels, to generate spatial embeddings for each frame. This process transforms raw video frames into compact, informative representations. The subsequent temporal modeling stage processes these sequential embeddings using a transformer-like structure consisting of one encoder and three decoders. Each encoder and decoder comprises 10 blocks, incorporating causal dilated convolutions to expand the receptive field while maintaining temporal causality. Together with the masked self-attention and cross-attention, the model effectively captures temporal dependencies in the surgical video, preserving the causal nature of the phase recognition task.

\subsubsection{Loss Function}
\label{losssection}
The loss function is a combination of classification loss $L_{cls}$ for each frame and smooth loss $L_{smo}$ \cite{farha2019ms}. The classification loss is a cross-entropy loss, while the smooth loss calculates
the mean squared error over the frame-wise probabilities. The final loss function $\mathcal{L}$ is:

\begin{equation}
\label{loss}
\mathcal{L} = \frac{1}{T} \sum_{t=1}^T \mathcal{H}(S(p_{y_t,t}), {y}_t) 
+ \lambda \frac{1}{T C} \sum_{t=2}^T \sum_{c=1}^C \text{clamp}\left( \Delta_t^2, 0, 16 \right),
\end{equation}
\begin{equation}
\Delta_t = \log \left( \text{Softmax}(p_{c,t}) \right) - \log \left( \text{Softmax}(p_{c,t-1}) \right),
\end{equation}

where \(\mathcal{H}(S(p_{y_t,t}), \hat{y}_t)\) is the standard cross-entropy between the softmax probabilities \(S(p_{y_t,t})\) for the predicted logits and the ground truth label \({y}_t\) at timestep \(t\). The softmax function is defined as \(\text{Softmax}(p_{c,t}) = \frac{\exp(p_{c,t})}{\sum_{c'=1}^C \exp(p_{c',t})}\), which converts the logits \(p_{c,t}\) into probabilities. The clamp function \(\text{clamp}(x, 0, 16)\) limits the value of \(x\) to the range \([0, 16]\) for temporal smoothness penalty. The hyperparameter \(\lambda\) controls the weight of the temporal smoothness loss relative to the classification loss, and was empirically set to 0.15 for all experiments reported in the results section. Finally, \(C\) denotes the number of classes, and \(T\) represents the total number of frames.

\subsubsection{Training Details}
We conducted our experiments on a GeForce RTX 2080 Ti GPU (NVIDIA Corp., CA, USA). In the first stage, we trained a ResNet feature extractor on individual frames using a learning rate of 1$\times$10$^{-5}$, cross-entropy loss, and a batch size of 32. After this stage, video features were extracted using the trained model and saved as feature representations with dimensions (number of video frames, 2048). In the second stage, we trained the temporal model exclusively on these saved video features, using a learning rate of 5$\times$10$^{-4}$ for 200 epochs, using the loss function described in \ref{losssection}.

\subsection{Evaluation Metrics}

For evaluation, we noted variations in the calculation approaches used in previous surgical phase recognition studies. \citeauthor{funke2023metrics} \cite{funke2023metrics} provided a structured overview of evaluation results on Cholec80 \cite{twinanda2016endonet} and AutoLaparo \cite{wang2022autolaparo}. To ensure consistency and comprehensiveness when evaluating all implemented models, we utilized the code base developed by \citeauthor{funke2023metrics} for all models in this study.

Accuracy is calculated at the video level as the percentage of correctly recognized frames across the entire video.
Precision, recall, and Jaccard are calculated for each phase individually and then averaged over all phases. The edit score \cite{lea2016segmental} quantifies the similarity of two sequences. It is based on the Levenshtein or edit distance and tallies the minimum number of insertions, deletions, and replacement operations required to convert one segment sequence into another. The F1 score or F1@$\tau$ \cite{lea2017temporal} compares the Intersection over Union (IoU) of each segment with respect to the corresponding ground truth based on some threshold $\tau$/100. Standard deviations are calculated across videos in the testing set.

\begin{table}[htbp]
\centering
\begin{threeparttable} 
\caption{Experimental results (\%) on RAMIE dataset (Mean ± Standard Deviation is computed across videos in the test set)}
\label{RAMIE-result1}
\begin{tabular}{lllll}
\toprule
 & Accuracy & Precision & Recall & Jaccard \\
\midrule
SV-RCNet & 75.42 ± 3.88 & 75.54 ± 4.00 & 70.12 ± 5.02 & 56.56 ± 5.55 \\
TeCNO & \textbf{78.46 ± 3.97} & 73.87 ± 4.60 & 73.56 ± 5.10 & 58.34 ± 4.75 \\
TMRNet & 72.86 ± 4.82 & 76.56 ± 6.04 & 57.12 ± 5.85 & 46.87 ± 5.42 \\
Trans-SVnet & 75.15 ± 4.09 & 74.79 ± 6.62 & 68.43 ± 5.98 & 55.25 ± 6.23 \\
Ours & 78.28 ± 4.42 & \textbf{77.28 ± 5.37} & \textbf{76.41 ± 6.01} & \textbf{61.94 ± 7.24} \\
\bottomrule
\end{tabular}
\end{threeparttable}
\end{table}

\begin{table}[htbp]
\centering
\begin{threeparttable} 
\caption{Experimental results (\%) on RAMIE dataset (Mean ± Standard Deviation is computed across videos in the test set)}
\label{RAMIE-result2}
\begin{tabular}{lllll}
\toprule
 & Edit Score & F1@25 & F1@50 & F1@75 \\
\midrule
SV-RCNet & 9.26 ± 1.39 & 11.94 ± 2.12 & 7.61 ± 1.43 & 4.04 ± 0.76 \\
TeCNO & 13.15 ± 1.83 & 17.79 ± 2.83 & 12.25 ± 2.86 & 6.52 ± 1.84 \\
TMRNet & 15.63 ± 1.96 & 19.34 ± 2.06 & 12.45 ± 1.27 & 5.55 ± 2.00 \\
Trans-SVnet & 6.85 ± 1.03 & 8.72 ± 1.32 & 5.63 ± 1.15 & 2.78 ± 0.75 \\
Ours & \textbf{59.50 ± 6.34} & \textbf{58.42 ± 4.45} & \textbf{45.08 ± 5.94} & \textbf{27.19 ± 3.41} \\
\bottomrule
\end{tabular}
\end{threeparttable}
\end{table}

\begin{figure}[ht]
    \centering
    \includegraphics[height=5.5cm]{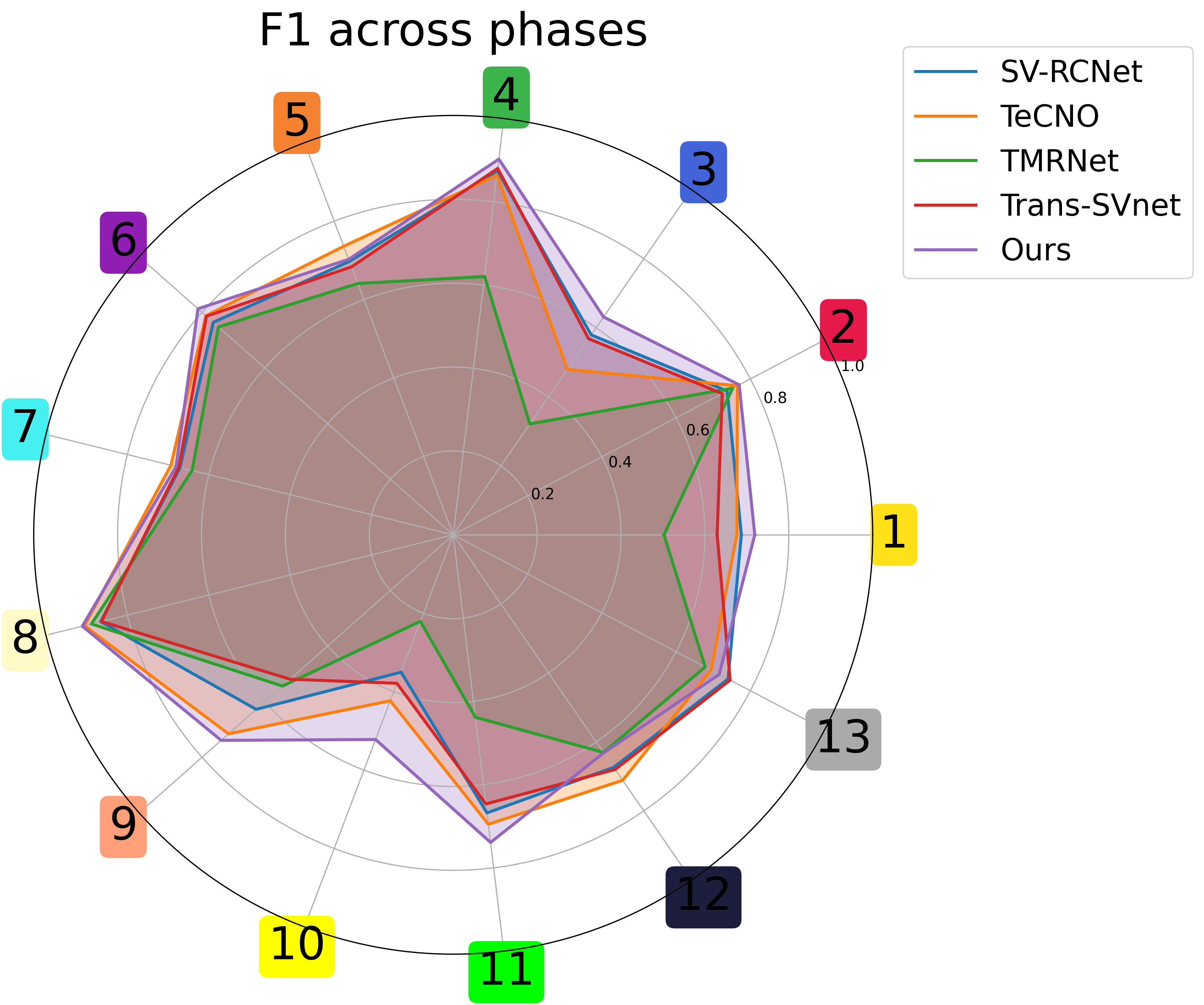}
    \hspace{0.0005\textwidth}
    \includegraphics[height=6cm]{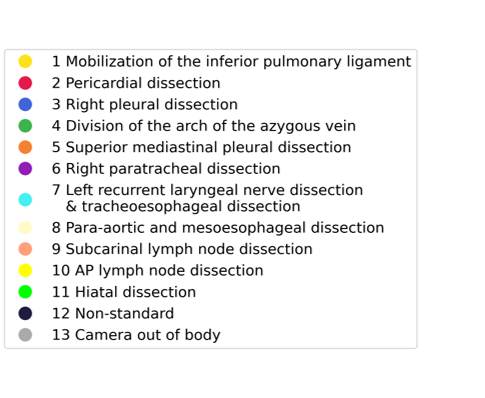}
    \caption{Mean F1 scores across surgical phases in RAMIE dataset}
    \label{radarwithlegend}
\end{figure}

\begin{figure}[htbp]
\raggedright 
\begin{tabular}{l} 
\includegraphics[height=5.5cm]{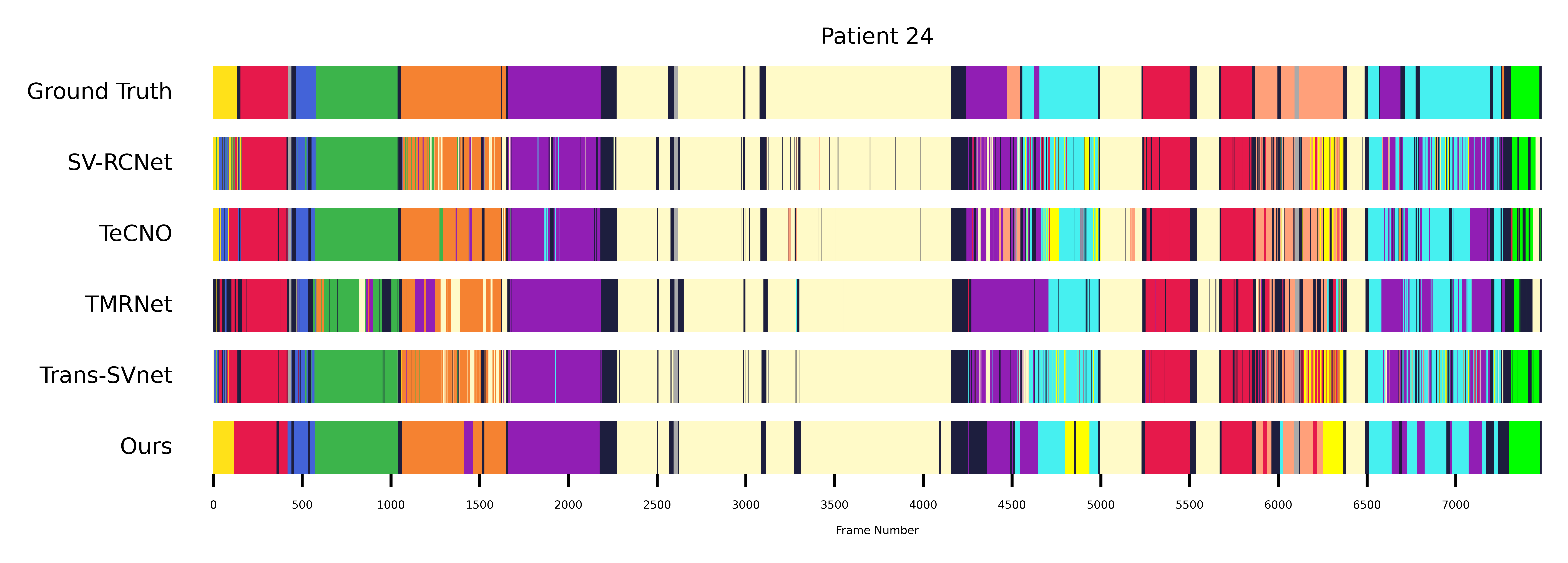}
\end{tabular}
\caption[Qualitative result]
{\label{QualitativeResult} Qualitative Result on RAMIE dataset}
\end{figure}

As shown in Table \ref{RAMIE-result1} and Table \ref{RAMIE-result2}, our model achieved improved performance across most metrics on our RAMIE dataset. Figure \ref{radarwithlegend} highlights that Phase 3 (Right pleural dissection) and Phase 10 (AP lymph node dissection) present the most significant challenges, both being relatively short surgical phases. Qualitative analysis revealed frequent misclassifications between Phase 10 (AP lymph node dissection) and two other phases: Phase 7 (Left laryngeal nerve dissection) and Phase 9 (Subcarinal dissection). From qualitative results similar to Figure \ref{QualitativeResult}, we observed that classification errors predominantly occur in proximity to phase transitions, suggesting that accurately delineating the boundaries between these phases remains a key challenge for the model. 

\subsection{Results on AutoLaparo dataset}
Table \ref{AutoLaparo} presents a comparison of the baseline models and our proposed model on the AutoLaparo dataset. While baseline results include only the mean of metrics, we provide both the mean and standard deviation across test set videos. Our model shows improved performance across all available metrics.

\begin{table}[!ht]
\centering
\caption{Performance comparison of different models on AutoLaparo dataset (baseline results from \cite{wang2022autolaparo})}
\label{AutoLaparo}
\begin{tabular}{lccccc}
\toprule
 & Accuracy & Precision & Recall & Jaccard \\
\midrule
SV-RCNet & 75.62 & 64.02 & 59.70 & 47.15 \\
TeCNO & 77.27 & 66.92 & 64.60 & 50.67  \\
TMRNet & 78.20 & 66.02 & 61.47 & 49.59 \\
Trans-SVnet & 78.29 & 64.21 & 62.11 & 50.65 \\
Ours & \textbf{83.18 ± 9.75} & \textbf{80.17 ± 12.12} & \textbf{77.05 ± 10.34} & \textbf{65.84 ± 12.59}\\
\bottomrule
\end{tabular}
\end{table}

The observed improvements in the performance of our model can be attributed to several key factors. The incorporation of a causal hierarchical attention mechanism within the encoder-decoder structure has a good ability to capture relevant temporal dependencies in complex sequences. The multi-layer architecture of the decoder allows iterative refinement of predictions through each decoder layer. In addition, the smoothing loss term is effective in addressing over-segmentation issues, where the model incorrectly divides continuous surgical phases into an excessive number of short and distinct segments. This subsequently led to higher scores for metrics that evaluate the temporal continuity of segments, such as the edit score and the F1 score with overlap.

\section{DISCUSSION}

The AutoLaparo and RAMIE datasets share the same task but differ significantly in their characteristics. The RAMIE dataset is notably more complex, featuring richer temporal dynamics and greater variability in phase sequences, further compounded by the intricate anatomical context of esophagectomy procedures. A robust temporal model must adapt to these diverse patterns, and our improved model demonstrates notable performance gains. However, over-segmentation remains a challenge, largely due to the inclusion of transition movements in the dataset. Additionally, the imbalance in phase lengths across patients, with varying phase sequences, poses difficulties, especially for less-represented phases. Further advancements in model development are needed to address these issues.

Evaluating surgical phase recognition in robot-assisted esophagectomy presents unique challenges, particularly in identifying phase transitions when key anatomical structures are not yet visible. Precise phase timing is critical for guiding the surgeon and preventing complications. However, current metrics fail to fully capture the models' ability to recognize phase beginnings, which is crucial for clinical applications. This aspect requires further exploration.

Additionally, surgical phases vary in risk, with some being more prone to complications and requiring greater recognition accuracy. Future work should prioritize improving accuracy for these critical phases. Multi-surgeon studies are essential for establishing clinically relevant benchmarks, which will enhance the translational potential of surgical phase recognition systems. Ensuring accuracy and adaptability across surgical practices is key to improving real-world utility in the operating room.

\section{CONCLUSIONS}
In conclusion, we have developed a new surgical phase recognition dataset specific to RAMIE, with the aim of making it publicly available in the future. Using this dataset, we conducted a comparative study of existing surgical phase recognition models on this data. Our newly developed model, which incorporates an encoder-decoder structure with causal hierarchical attention for temporal modelling demonstrates superior performance.
The results provide valuable insights into overall model performance as well as performance on specific surgical phases. Qualitative analysis has revealed challenges such as over-segmentation and specific error patterns, highlighting areas for future improvement.

This work establishes a foundation for advancing surgical phase recognition models in RAMIE. By addressing the identified challenges, we aim to improve the reliability and clinical applicability of automated surgical phase recognition systems, potentially enhancing surgical outcomes and patient safety in robot-assisted esophagectomy.

\acknowledgments 
 
This research was funded by Stichting Hanarth
Fonds, study number: 2022-13. It is part of the INTRA-SURGE (INTelligent computeR-Aided Surgical gUidance for Robot-assisted
surGEry) project aimed at advancing the future of surgery.

\bibliography{main} 

\begin{thebibliography}{23}
\providecommand{\natexlab}[1]{#1}
\providecommand{\url}[1]{\texttt{#1}}
\expandafter\ifx\csname urlstyle\endcsname\relax
  \providecommand{\doi}[1]{doi: #1}\else
  \providecommand{\doi}{doi: \begingroup \urlstyle{rm}\Url}\fi

\bibitem[Bray et~al.(2024)Bray, Laversanne, Sung, Ferlay, Siegel, Soerjomataram, and Jemal]{bray2024global}
Freddie Bray, Mathieu Laversanne, Hyuna Sung, Jacques Ferlay, Rebecca~L Siegel, Isabelle Soerjomataram, and Ahmedin Jemal.
\newblock Global cancer statistics 2022: Globocan estimates of incidence and mortality worldwide for 36 cancers in 185 countries.
\newblock \emph{CA: a cancer journal for clinicians}, 74\penalty0 (3):\penalty0 229--263, 2024.

\bibitem[Fuchs et~al.(2022)Fuchs, Collins, Babic, DuCoin, Meireles, Grimminger, Read, Abbas, Sallum, M{\"u}ller-Stich, et~al.]{fuchs2022robotic}
Hans~F Fuchs, Justin~W Collins, Benjamin Babic, Christopher DuCoin, Ozanan~R Meireles, Peter~P Grimminger, Matthew Read, Abbas Abbas, Rubens Sallum, Beat~P M{\"u}ller-Stich, et~al.
\newblock Robotic-assisted minimally invasive esophagectomy (ramie) for esophageal cancer training curriculum—a worldwide delphi consensus study.
\newblock \emph{Diseases of the Esophagus}, 35\penalty0 (6):\penalty0 doab055, 2022.

\bibitem[van~der Sluis et~al.(2018)van~der Sluis, Ruurda, van~der Horst, Goense, and van Hillegersberg]{van2018learning}
Pieter~C van~der Sluis, Jelle~P Ruurda, Sylvia van~der Horst, Lucas Goense, and Richard van Hillegersberg.
\newblock Learning curve for robot-assisted minimally invasive thoracoscopic esophagectomy: results from 312 cases.
\newblock \emph{The Annals of Thoracic Surgery}, 106\penalty0 (1):\penalty0 264--271, 2018.

\bibitem[Den~Boer et~al.(2023)Den~Boer, Jaspers, De~Jongh, Pluim, Van Der~Sommen, Boers, van Hillegersberg, Van~Eijnatten, and Ruurda]{den2023deep}
RB~Den~Boer, TJM Jaspers, C~De~Jongh, JPW Pluim, F~Van Der~Sommen, T~Boers, R~van Hillegersberg, MAJM Van~Eijnatten, and JP~Ruurda.
\newblock Deep learning-based recognition of key anatomical structures during robot-assisted minimally invasive esophagectomy.
\newblock \emph{Surgical endoscopy}, 37\penalty0 (7):\penalty0 5164--5175, 2023.

\bibitem[Sato et~al.(2022)Sato, Fujita, Matsuzaki, Takeshita, Fujiwara, Mitsunaga, Kojima, Mori, and Daiko]{sato2022real}
Kazuma Sato, Takeo Fujita, Hiroki Matsuzaki, Nobuyoshi Takeshita, Hisashi Fujiwara, Shuichi Mitsunaga, Takashi Kojima, Kensaku Mori, and Hiroyuki Daiko.
\newblock Real-time detection of the recurrent laryngeal nerve in thoracoscopic esophagectomy using artificial intelligence.
\newblock \emph{Surgical Endoscopy}, 36\penalty0 (7):\penalty0 5531--5539, 2022.

\bibitem[Takeuchi et~al.(2022)Takeuchi, Kawakubo, Saito, Maeda, Matsuda, Fukuda, Nakamura, and Kitagawa]{takeuchi2022automated}
Masashi Takeuchi, Hirofumi Kawakubo, Kosuke Saito, Yusuke Maeda, Satoru Matsuda, Kazumasa Fukuda, Rieko Nakamura, and Yuko Kitagawa.
\newblock Automated surgical-phase recognition for robot-assisted minimally invasive esophagectomy using artificial intelligence.
\newblock \emph{Annals of Surgical Oncology}, 29\penalty0 (11):\penalty0 6847--6855, 2022.

\bibitem[Czempiel et~al.(2020)Czempiel, Paschali, Keicher, Simson, Feussner, Kim, and Navab]{czempiel2020tecno}
Tobias Czempiel, Magdalini Paschali, Matthias Keicher, Walter Simson, Hubertus Feussner, Seong~Tae Kim, and Nassir Navab.
\newblock Tecno: Surgical phase recognition with multi-stage temporal convolutional networks.
\newblock In \emph{Medical Image Computing and Computer Assisted Intervention--MICCAI 2020: 23rd International Conference, Lima, Peru, October 4--8, 2020, Proceedings, Part III 23}, pages 343--352. Springer, 2020.

\bibitem[Brandenburg et~al.(2023)Brandenburg, Jenke, Stern, Daum, Schulze, Younis, Petrynowski, Davitashvili, Vanat, Bhasker, et~al.]{brandenburg2023active}
Johanna~M Brandenburg, Alexander~C Jenke, Antonia Stern, Marie~TJ Daum, Andr{\'e} Schulze, Rayan Younis, Philipp Petrynowski, Tornike Davitashvili, Vincent Vanat, Nithya Bhasker, et~al.
\newblock Active learning for extracting surgomic features in robot-assisted minimally invasive esophagectomy: a prospective annotation study.
\newblock \emph{Surgical Endoscopy}, 37\penalty0 (11):\penalty0 8577--8593, 2023.

\bibitem[Maier-Hein et~al.(2020)Maier-Hein, Eisenmann, Sarikaya, M{\"a}rz, Collins, Malpani, Fallert, Feussner, Giannarou, Mascagni, et~al.]{maier2020surgical}
Lena Maier-Hein, Matthias Eisenmann, Duygu Sarikaya, Keno M{\"a}rz, Toby Collins, Anand Malpani, Johannes Fallert, Hubertus Feussner, Stamatia Giannarou, Pietro Mascagni, et~al.
\newblock Surgical data science-from concepts to clinical translation.
\newblock \emph{arXiv preprint arXiv:2011.02284}, 2, 2020.

\bibitem[van Boxel et~al.(2019)van Boxel, van Hillegersberg, and Ruurda]{van2019outcomes}
Gijsbert van Boxel, Richard van Hillegersberg, and Jelle Ruurda.
\newblock Outcomes and complications after robot-assisted minimally invasive esophagectomy.
\newblock \emph{Journal of Visualized Surgery}, 5, 2019.

\bibitem[Kingma et~al.(2020)Kingma, Read, Van~Hillegersberg, Chao, and Ruurda]{kingma2020standardized}
BF~Kingma, M~Read, R~Van~Hillegersberg, YK~Chao, and JP~Ruurda.
\newblock A standardized approach for the thoracic dissection in robotic-assisted minimally invasive esophagectomy (ramie).
\newblock \emph{Diseases of the Esophagus}, 33\penalty0 (Supplement\_2):\penalty0 doaa066, 2020.

\bibitem[Wang et~al.(2022)Wang, Lu, Long, Zhong, Cheung, Dou, and Liu]{wang2022autolaparo}
Ziyi Wang, Bo~Lu, Yonghao Long, Fangxun Zhong, Tak-Hong Cheung, Qi~Dou, and Yunhui Liu.
\newblock Autolaparo: A new dataset of integrated multi-tasks for image-guided surgical automation in laparoscopic hysterectomy.
\newblock In \emph{International Conference on Medical Image Computing and Computer-Assisted Intervention}, pages 486--496. Springer, 2022.

\bibitem[Jin et~al.(2017)Jin, Dou, Chen, Yu, Qin, Fu, and Heng]{jin2017sv}
Yueming Jin, Qi~Dou, Hao Chen, Lequan Yu, Jing Qin, Chi-Wing Fu, and Pheng-Ann Heng.
\newblock Sv-rcnet: workflow recognition from surgical videos using recurrent convolutional network.
\newblock \emph{IEEE transactions on medical imaging}, 37\penalty0 (5):\penalty0 1114--1126, 2017.

\bibitem[Jin et~al.(2021)Jin, Long, Chen, Zhao, Dou, and Heng]{jin2021temporal}
Yueming Jin, Yonghao Long, Cheng Chen, Zixu Zhao, Qi~Dou, and Pheng-Ann Heng.
\newblock Temporal memory relation network for workflow recognition from surgical video.
\newblock \emph{IEEE Transactions on Medical Imaging}, 40\penalty0 (7):\penalty0 1911--1923, 2021.

\bibitem[Gao et~al.(2021)Gao, Jin, Long, Dou, and Heng]{gao2021trans}
Xiaojie Gao, Yueming Jin, Yonghao Long, Qi~Dou, and Pheng-Ann Heng.
\newblock Trans-svnet: Accurate phase recognition from surgical videos via hybrid embedding aggregation transformer.
\newblock In \emph{Medical Image Computing and Computer Assisted Intervention--MICCAI 2021: 24th International Conference, Strasbourg, France, September 27--October 1, 2021, Proceedings, Part IV 24}, pages 593--603. Springer, 2021.

\bibitem[Twinanda et~al.(2016)Twinanda, Shehata, Mutter, Marescaux, De~Mathelin, and Padoy]{twinanda2016endonet}
Andru~P Twinanda, Sherif Shehata, Didier Mutter, Jacques Marescaux, Michel De~Mathelin, and Nicolas Padoy.
\newblock Endonet: a deep architecture for recognition tasks on laparoscopic videos.
\newblock \emph{IEEE transactions on medical imaging}, 36\penalty0 (1):\penalty0 86--97, 2016.

\bibitem[Yi et~al.(2021)Yi, Wen, and Jiang]{yi2021asformer}
Fangqiu Yi, Hongyu Wen, and Tingting Jiang.
\newblock Asformer: Transformer for action segmentation.
\newblock \emph{arXiv preprint arXiv:2110.08568}, 2021.

\bibitem[Kuehne et~al.(2014)Kuehne, Arslan, and Serre]{Kuehne12}
H.~Kuehne, A.~B. Arslan, and T.~Serre.
\newblock The language of actions: Recovering the syntax and semantics of goal-directed human activities.
\newblock In \emph{Proceedings of Computer Vision and Pattern Recognition Conference (CVPR)}, 2014.

\bibitem[Stein and McKenna(2013)]{stein2013combining}
Sebastian Stein and Stephen~J McKenna.
\newblock Combining embedded accelerometers with computer vision for recognizing food preparation activities.
\newblock In \emph{Proceedings of the 2013 ACM international joint conference on Pervasive and ubiquitous computing}, pages 729--738, 2013.

\bibitem[Farha and Gall(2019)]{farha2019ms}
Yazan~Abu Farha and Jurgen Gall.
\newblock Ms-tcn: Multi-stage temporal convolutional network for action segmentation.
\newblock In \emph{Proceedings of the IEEE/CVF conference on computer vision and pattern recognition}, pages 3575--3584, 2019.

\bibitem[Funke et~al.(2023)Funke, Rivoir, and Speidel]{funke2023metrics}
Isabel Funke, Dominik Rivoir, and Stefanie Speidel.
\newblock Metrics matter in surgical phase recognition.
\newblock \emph{arXiv preprint arXiv:2305.13961}, 2023.

\bibitem[Lea et~al.(2016)Lea, Reiter, Vidal, and Hager]{lea2016segmental}
Colin Lea, Austin Reiter, Ren{\'e} Vidal, and Gregory~D Hager.
\newblock Segmental spatiotemporal cnns for fine-grained action segmentation.
\newblock In \emph{Computer Vision--ECCV 2016: 14th European Conference, Amsterdam, The Netherlands, October 11-14, 2016, Proceedings, Part III 14}, pages 36--52. Springer, 2016.

\bibitem[Lea et~al.(2017)Lea, Flynn, Vidal, Reiter, and Hager]{lea2017temporal}
Colin Lea, Michael~D Flynn, Rene Vidal, Austin Reiter, and Gregory~D Hager.
\newblock Temporal convolutional networks for action segmentation and detection.
\newblock In \emph{proceedings of the IEEE Conference on Computer Vision and Pattern Recognition}, pages 156--165, 2017.

\end{thebibliography}

\end{document}